\newcommand{\RNum}[1]{\uppercase\expandafter{\romannumeral #1\relax}}

\documentclass{IEEEtran4PSCC}
\ifCLASSINFOpdf
\else
\fi
\usepackage[cmex10]{amsmath}

\usepackage{amsfonts}
\usepackage{multirow}
\usepackage{booktabs,colortbl}
\usepackage{makecell}

\usepackage{siunitx}
\usepackage{subcaption}
\usepackage{graphicx}

\DeclareMathOperator*{\argmin}{arg\,min}

\begin{document}
	%
	\title{Unit Commitment using Nearest Neighbor as a Short-Term Proxy}

	
	
	%
	\author{\IEEEauthorblockN{Gal Dalal\IEEEauthorrefmark{1},
			Elad Gilboa\IEEEauthorrefmark{1},
			Shie Mannor\IEEEauthorrefmark{1}, and
			Louis Wehenkel\IEEEauthorrefmark{2}}
		\IEEEauthorblockA{\IEEEauthorrefmark{1}Department of Electrical Engineering \\
			Technion, Israel Institute of Technology,
			\{gald,egilboa\}@tx.technion.ac.il, shie@ee.technion.ac.il}
		\IEEEauthorblockA{\IEEEauthorrefmark{2}Montefiore Institute -- Department of Electrical Engineering and Computer Science\\
			University of Li\`ege, L.Wehenkel@ulg.ac.be\\
		}
	}


	\maketitle
	
	\begin{abstract}
		We devise the Unit Commitment Nearest Neighbor (UCNN) algorithm to be used as a proxy for quickly approximating outcomes of short-term decisions, to make tractable hierarchical long-term assessment and planning for large power systems. Experimental results on updated versions of IEEE-RTS79 and IEEE-RTS96 show high accuracy measured on operational cost, achieved in runtimes that are lower in several orders of magnitude than the traditional approach.
	\end{abstract}
	

	%
	\IEEEpeerreviewmaketitle

	\section{Introduction}
	
	Unit commitment (UC) is solved daily by Transmission System Operators (TSO) worldwide as part of the market clearing process, to ensure safe operation. 
	Typically, the resulting mathematical problem is either a deterministic or stochastic Mixed Integer-Linear Program (MILP). It is solved accurately for the following day, taking into account all available information on generation and demand, along with exogenous factors such as renewable generation forecast.  
	
	As intermittent generation capacity  is increasing regularly in recent years, more stochasticity is involved in power system operation, affecting the way planning is done not only in the day-ahead time horizon but in all different time horizons \cite{GARPUR_2.2,parsons2004grid,milligan2000modelling}. The complex dependence between the different time-horizons and the high uncertainty in long time-horizons  makes long-term planning challenging. As demonstrated in \cite{dalal2016distributed}, solving an extensive amount of UC problems to mimic short-term decision-making does not scale well to realistic grids, with thousands of nodes, generators and loads. This is especially burdensome in planning for horizons of months to years, such as scheduling outages for asset maintenance. This mid- to long-term planning problem necessitates consideration of shorter timescale operation.	Outage scheduling, for instance, needs to be coordinated with short-term operation, namely the TSO's intervention in the day-ahead market clearing. For brevity, from this point on, we jointly refer both to mid- and long-term time horizons as long-term.
	
	Planning under uncertainty is often done using stochastic optimization. This involves generation of scenarios, which in the case of long-term planning span over months or even dozens of years. In this context, scenario evolution is dependent on the sought plan and contains daily and hourly states of the system and exogenous conditions such as wind generation and consumption. To illustrate this, consider a maintenance planner, assessing several alternatives for next year's proposed outage schedule. To evaluate each of the schedules, he needs to examine the network's ability to comply with security constrained UC during the proposed outages in the schedules. He will thus reproduce different possible network conditions during each of the year's months, in terms of likely nodal wind generation and demand during that month. For each of the reproduced conditions, a UC problem will be solved given the specific future topology of the grid under the outages planned for this month. The planner will conduct this using simulation, iterating many times for each of the year's months, per each of the optional outage schedules. The more accurate he wishes the result to be, the more wind and demand samples he should feed to his UC solver. Each resulting UC solution can be used to evaluate the outage schedule in multiple ways: counting the number of feasible UC programs; averaging UC cost; averaging load lost amount; used as a reference for calculating  costs in finer-grained hourly simulation, such as re-dispatch and re-commitment of generators. 
	
	Motivated by the above use-case, in this work we consider the need to solve numerous UC problem instances, for which the solution accuracy is not of the first priority.
	
	\subsection{Contribution}
	Our claim is that in large networks, in the context of long-term planning, approximated \emph{proxy} methods are necessary for assessment of cost and reliability. We thus introduce the following concept -- learning a proxy for approximating short-term decision-making outcomes, relieving the dependence of long-term assessment on accurate short-term simulations; ergo, allowing for a tractable assessment methodology. We use a well-known machine learning algorithm -- \emph{nearest neighbor classification} \cite{cover1967nearest}. Therefore, we call it UCNN.
	
	The methodology relies on a simple concept -- creating a large and diverse dataset that contains samples of the environment and grid conditions along with their respective UC solution. Consequently, during the assessment of an outage schedule, instead of solving the multiple UC problem instances required to simulate decisions taken, we merely choose among the already pre-computed UC solutions. The UC solution chosen to be used is the one with the closest conditions to the environment and grid conditions of the current UC problem needed to be solved; hence the phrase nearest neighbor.
	
	The essence of this method's advantage lies in the fact that planning and assessment for long horizons in stochastic environments require obtaining multiple samples (UC solutions), and the assumption that similar repetitive UC solutions will result in similar outputs (cost, reliability, etc.). Therefore, instead of repeating the expensive process of obtaining these samples (solving MILPs) for environment and grid conditions that often are repetitive within a single scenario and across different scenarios, utilize samples created ex-ante as representatives of sets of similar repetitive conditions. The initial creation of the dataset is a slow process which can either be done offline, or online by continually adding new solutions during the long-horizon assessment process itself.
	After obtaining the training set, UCNN reduces computation time in several orders of magnitude, with relatively little compromise in quality, as shown in Section \ref{sec:UCNN_experiments}. 
	Without this significant reduction in computation time, long-term assessment processes, which account for short-term decisions based on multiple UC instances, are deemed to be computationally intractable.

	
	\subsection{Related work on supervised learning}
	The literature contains several works that use machine learning for predicting outcomes of decision processes in power networks based on pre-solved various input conditions. Such methods are often under the category of supervised learning algorithms. We limit our survey to the problem of generation (re)scheduling and reserve activation. In \cite{cornelusse2007automatic}, frequency and active power time series were used for determining whether generator reserve activation is satisfactory or not. In this application, manual labeling of the data is required by experts, and there are only four possible label classes.
	Reliability is maximized in \cite{dalal2016hierarchical} by learning a function that assesses the implications of rescheduling. The sought output is a policy, that dynamically maps system states to rescheduling actions.
	In \cite{cornelusse2009supervised}, supervised learning was used for finding recourse strategies in generation management, by generating a training set via Monte-Carlo simulation of load and generation disturbances and then learning a  near-optimal recourse strategy to handle similar disturbances observed in real-time. 
	Recourse strategy learning was also investigated in \cite{rachelson2010combining}, where boosting is used to create binary classifiers for boolean variables of the mixed integer programs resulting from daily generation re-planning problems. 
	
	\section{Unit commitment nearest neighbor classification}
	We begin with defining the accurate UC solution; notations are adopted from \cite{dalal2016distributed}. The optimal UC decision $u_p^*(y_s,u_m)$ is the solution of the following optimization problem:
	\begin{equation} \label{eq:optimals}
	u_p^*(y_s,u_m) = \argmin_{u_p \in \mathcal{U}_p(u_m)} \quad  C_p(y_s,u_m,u_p),
	\end{equation}  
	where $u_m$ is a long-term planning decision that dictates the topology of the  network (e.g., outages in transmission lines 2 and 5); $\mathcal{U}_p(u_m)$ is the set of feasible UC schedules with respect to $u_m$;  $C_p$ is the overall UC cost, consisting of generation, start-up, wind curtailment, and load shedding costs; $y_s$ is the day-ahead forecast of hourly nodal renewable generation and demand. We additionally denote the optimal value of the objective function in \eqref{eq:optimals} by
	\begin{equation} \label{eq:optimals2}
	C_p^*(y_s,u_m) \equiv  C_p(y_s,u_m,u_p^*(y_s,u_m)).
	\end{equation}

	In this work, we use the DC power-flow formulation constrained to 
	available generator capacities, minimum up/down times, ramp-rates, line flow constraints, and N-1 security constraints (for some of the simulations).
	It also includes wind curtailment and load shedding decision variables. This results in a MILP that can be solved using commercial solvers \cite{CPLEX}. The exact UC formulation is given in Appendix~\ref{sec:unit_commitment}.  For simplicity, we only consider transmission line outages, albeit our method is not limited to them. By repeating the same steps described in this paper for additional types of outages, one can also account for maintenance of equipment such as generators, transformers, shunt elements, etc.
	
	 The UC problem we solve is a single-stage formulation that comes to estimate the market-clearing process. In the real world, two-stage UC problems are often being solved (usually in Europe), in a continental scale at a first stage, and in a zonal level at a later stage. Notice this day-ahead  problem does not probabilistically account for possible real-time balancing market realizations. Our formulation, often referred to as \emph{inefficient market} \cite{pineda2016impact}, is deterministic given forecast value $y_s$. This is in accordance with the purpose our UC proxy is serving: estimate long probabilistic paths based on multiple deterministic day-ahead solutions, which serve either as samples for higher-level statistics or as a baseline for finer-grained hourly simulation such as in \cite{dalal2016distributed}.
	Next, we introduce the proposed supervised learning approach to solve this problem.
	
	\subsection{Supervised learning algorithm}
	One definition for supervised learning is the problem of predicting some output value (label $l$)	
	given some input (sample $x$), while having on access to a \emph{training set} $\{(x^j,l^j)\}_{j=1}^{n}$ composed of $n$ input-output pairs.  In this work, $x=(y_s,u_m)$; i.e., it denotes the inputs to a UC optimization problem, and $l=(u_p^*(y_s,u_m), C_p^*(y_s,u_m))$; i.e., it denotes the pair of UC solution for $x$ and its cost.
	
	We now show how to utilize a well-known and popular classification algorithm -- nearest neighbor (NN) \cite{bhatia2010survey} -- to construct a proxy that replaces a computationally expensive MILP solver with an approximate solver and thus allow for computation time that is several orders of magnitude lower. Instead of finding exact solutions to \eqref{eq:optimals}, our solver finds high-quality approximate solutions to it, denoted by $(\hat{u}_p^*(y_s,u_m), \hat{C}_p^*(y_s,u_m))$. It does so by solving a much less complex problem: finding the most similar conditions to the environment and grid conditions $(y_s,u_m)$, out of the ones stored in the training set. The family of NN algorithms was shown to work well on a large variety of problems \cite{bhatia2010survey}. We choose it, since it is in principle able to make consistent predictions over very complex output spaces (in our case a set of pairs $(u^{*}_{p}, C^{*}_{p})$) provided that $n$ is large enough.
	
	\begin{figure*}[h]
		\centering
		\includegraphics[scale=0.41]{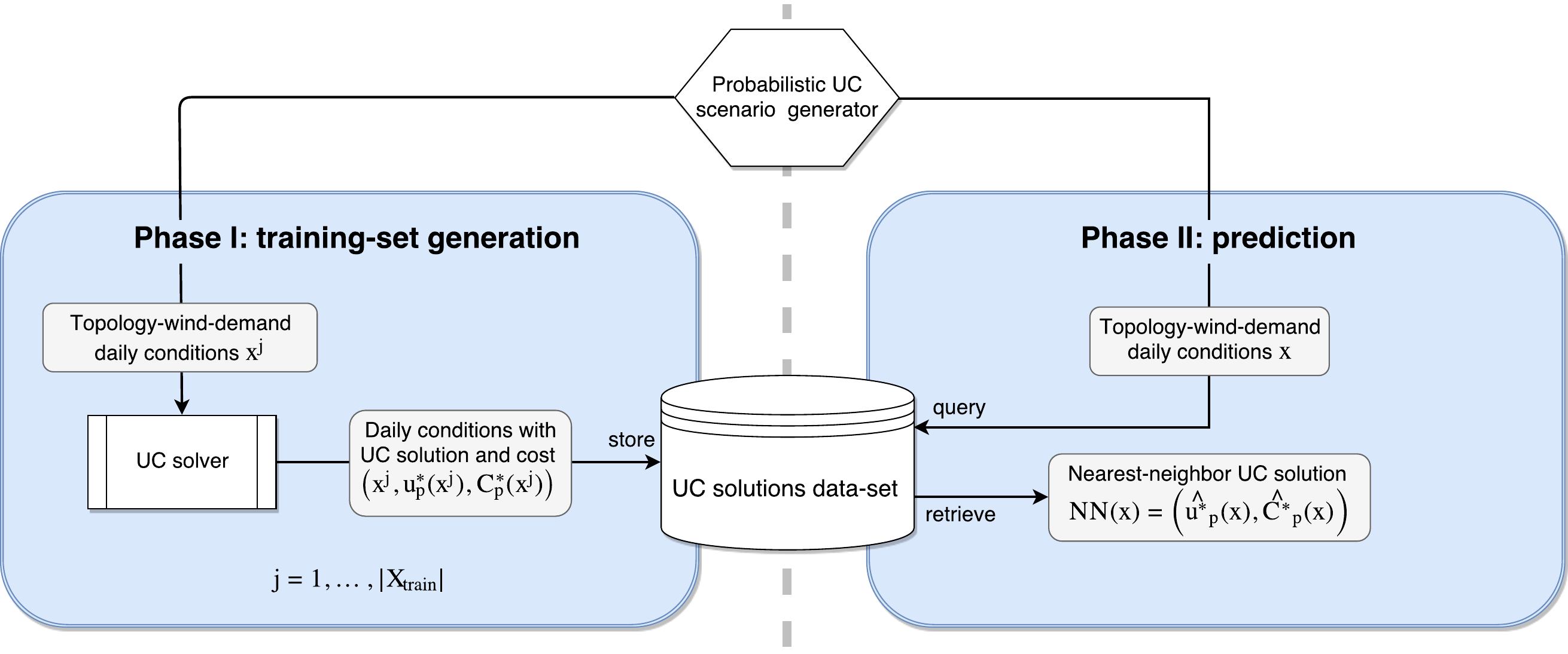}
		\caption{UCNN algorithm diagram. In an initial phase, multiple UC scenarios are generated and solved. Each such scenario is then stored along with its solution and cost to create a large and diverse dataset, also referred to as \emph{training set}. In a second phase, when a new UC problem instance is received, an approximate UC solution is obtained by finding the nearest neighbor among the existing solutions in the dataset, i.e., a pre-solved similar problem instance. This is to replace the usage of the computationally expensive UC solver.}
		\label{fig:block diagram}
	\end{figure*}

	\subsection{Phase \RNum{1}: training set generation}
	For the application of supervised learning we first need to build a large dataset of pre-solved UC problems, a process which we refer to as \emph{training set generation}. It involves an initial generation of a set of inputs $\{x^{j} = (y^{j}_s,u^{j}_m)\}_{j=1}^{n}$, drawn from the marginal distribution expected to be used during the long-term planning process, and solving each of them accurately; i.e., obtain its output label $l^{j}= (u_p^*(y^{j}_s,u^{j}_m), C_p^*(y^{j}_s,u^{j}_m))$  as defined in \eqref{eq:optimals2} by solving \eqref{eq:optimals}.

	\subsection{Definition of features and distance measure}
	Before describing the prediction phase, we first need to set a distance measure for quantifying similarity for choosing nearest neighbors. For that, we define a feature function: a mapping from the original inputs $x$ to a vector $$\phi(x) = [D_{\text{d.a}}(x);W_{\text{d.a}}(x);\text{top}_{\text{d.a}}(x)],$$ 
	where $D_{\text{d.a}}(x) \in \mathbb{R}^{24\times n_b}$ is a 24-hour demand forecast for the $n_b$ buses; $W_{\text{d.a}}(x) \in \mathbb{R}^{24\times n_w}$ is a 24-hour wind generation forecast for the $n_w$ wind generators; and $\text{top}_{\text{d.a}}(x) \in \{0,1\}^{n_l}$ is a daily network topology of the $n_l$ transmission lines: element $i$ in $\text{top}_{\text{d.a}}$ is $0$ if line $i$ is offline, and $1$ otherwise. The three variables above are flattened and concatenated to form a single column.
	
	Using this representation, we measure the distance $d(x,x')$ between two UC daily conditions $x,x'$ via the $\xi$-weighted $L_2$-norm of their feature difference:
	\begin{equation} \label{eq:distance}
	||\phi(x)-\phi(x')||_\xi = \left[\sum_{i=1}^{\text{length}(\phi(x))}\xi_i^2(\phi_{i}(x)_-\phi_{i}(x'))^2\right]^{\frac{1}{2}}.
	\end{equation}
	
	The weights $\xi$ are used for expressing the importance of different entries in choosing the nearest neighbor. In our simulations, the $\xi_i$s multiplying the entries of $\text{top}_{\text{d.a}}$ are chosen to be $100$, whereas the rest are set to $1$. This choice reflects our belief that in terms of similarity, network topology is more relevant than demand and wind forecast. In addition,  $\xi$ is used for scaling different units, i.e., for comparing binary values stemming from $\text{top}_{\text{d.a}}$ and [MW] values stemming from $D_{\text{d.a}}^{\text{cs}},~W_{\text{d.a}}^{\text{cs}}$. Further research in the field of feature selection and \emph{metric learning} \cite{bellet2013survey} is anticipated by the authors in this context.
	
	\subsection{Phase \RNum{2}: prediction}	
	
	The prediction of UC solution and cost $\hat{l}(x) = (\hat{u}_p^*(x), \hat{C}_p^*(x))$ for a new sample $x=(y_s,u_m)$ is done by first finding the index of the sample closest to it in the training set, i.e., by computing
	$${\text{NN}}(x) = \argmin_{j  \in \{1, \ldots, n\}} ||\phi(x)-\phi(x^{j})||_\xi;$$
	and then by setting 
	$$(\hat{u}_p^*(x), \hat{C}_p^*(x)) = (u_p^*(x^{{\text{NN}(x)}}), C_p^*(x^{{\text{NN}(x)}})) =  l^{{\text{NN}}(x)}.$$
		
	The complete UCNN flow is visualized in Fig.~\ref{fig:block diagram}.


	
	\section{Experiments} \label{sec:UCNN_experiments}
	
	We begin with presenting the indicators used in our evaluation methodology to test the merits of UCNN and then present our experimental results.
	\subsection{Evaluation methodology}
	\label{sec:evaluation methodolohy}
	From a machine learning perspective, the problem introduced is not a standard classification problem, where an algorithm is assessed by its probability to classify samples correctly. In our case, classifying a sample means choosing the optimal UC schedule for it and at the same time predicting the corresponding cost. There is, however, no obvious technique for comparing two UC schedules. They are represented as binary matrices that can be very different in terms of standard metrics, such as Manhattan distance; and yet practically identical in terms of operation, depending on the network test-case and choices such as the component ordering. Our setting thus necessitates a non-standard evaluation methodology.
	
	In light of the above, to assess our UCNN algorithm, we evaluate the cost prediction accuracy for samples taken from a test set ${\mathcal{X}}_{\text{test}}$, which is disjoint of the training set but generated in similar fashion. Let us, for brevity, abuse notations and denote by $C^{*}_p(x)$ the ``exact'' optimal cost in \eqref{eq:optimals2} for daily conditions $x$. Then, per each sample $x_\text{test} \in {\mathcal{X}}_{\text{test}}$, the approximate UC solution $\hat{u}_p^*(x_\text{test})$ is compared to its accurate counterpart $u_p^*(x_\text{test})$ via two accuracy measures: 
	\begin{enumerate}
		\item relative error $\frac{|\hat{C}^{*}_p(x_\text{test}) - C^{*}_p(x_\text{test})|}{C^{*}_p(x_\text{test})}$ ; and 
		\item linear correlation of $\hat{C}^{*}_p(x_\text{test})$ and $C^{*}_p(x_\text{test})$.
	\end{enumerate}
	These two measures are averaged over all samples in ${\mathcal{X}}_{\text{test}}$.

	\subsection{Experimental results}
	We run our experiments on a Sun cluster with several Intel(R) Xeon(R) servers  $@2.53$GHz, containing a total of 100 cores, each with 2GB of memory. 
	All code is written in Matlab \cite{matlab}. We use YALMIP \cite{YALMIP} to model the UC formulation and solve it using CPLEX \cite{CPLEX}.
	
\begin{figure} 
	 	\includegraphics[scale=0.45]{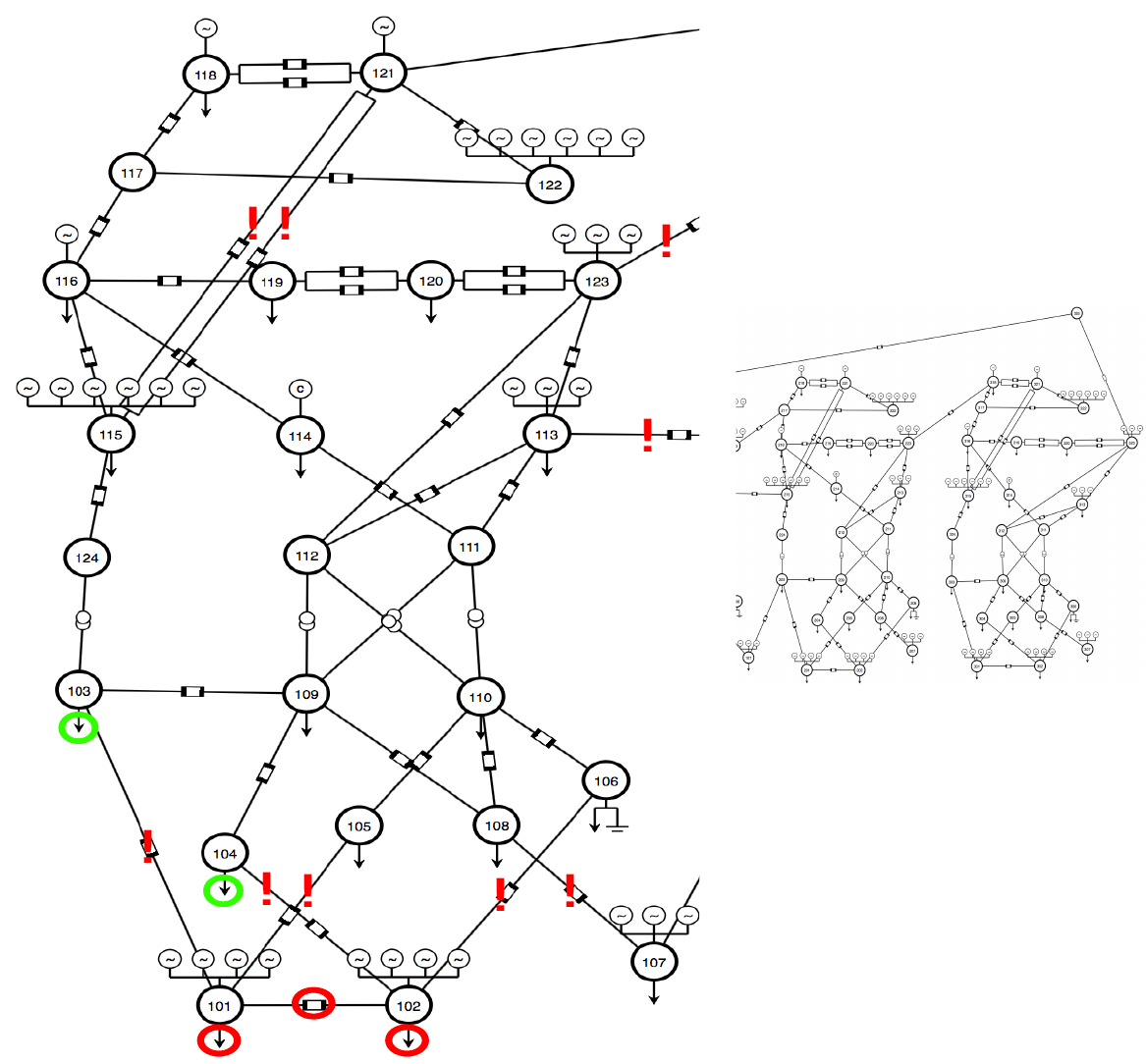}
	 	\caption{Modifications and candidate planned outages marking for the IEEE-RTS96 test-case. Red circles denote removal, green circles denote increase, and red exclamation marks denote candidate planned outage. The marking is presented on a single zone (enlarged) out of the three; exactly the same modifications are replicated to the other two zones as well.}
	 	\label{fig:case96_modifications}
	 \end{figure}

	For our simulation we use the IEEE-RTS79 and RTS96 test-cases; however, for compactness, in this section we only show the results for the larger of the two, RTS96. For more details please see Subsection~\ref{sec:further}.  We adopt updated generator parameters from Kirschen et. al \cite{pandzic2013comparison}, namely their capacities, minimum outputs, ramp up/down limits, minimum up/down times, price curves and start-up costs. Wind generation capacities and daily trajectories are based on real historical records from the US as published in \cite{UW_website}. Peak loads and daily demand profile are based on real data, taken from \cite{UW_website}. 
	 Value of lost load is set to $VOLL=1000[\frac{\$}{MWh}]$, taken from \cite{dvorkin2015hybrid} and wind-curtailment price is set to $C_{WC}=100[\frac{\$}{MWh}]$, taken from \cite{loisel2010valuation}. In addition, we slightly modify the test-case so as to create several 'bottleneck' areas to provide conditions for a variant set of  UC costs with relatively short simulation time. Per each of the three RTS96 zones, these modifications include i) removal of transmission line between bus 1 and 2; and ii) shift of loads from buses 1 and 2 to buses 3 and 4, respectively. The considered outages in each zone are in transmission lines with ID 2,3,4,5,11,25,26. Visualization is found in Fig. \ref{fig:case96_modifications}.
	 
	Training and test sets of labeled (solved) UC schedules are of sizes $|{\mathcal{X}}_{\text{train}}| = 14K$ , $|{\mathcal{X}}_{\text{test}}| = 1K$. The three components of each sample, i.e., $D_{\text{d.a}}(x)$, $W_{\text{d.a}}(x)$, and $\text{top}_{\text{d.a}}(x)$, are drawn independently and then concatenated into a single vector. 
	Both demand and wind processes are sampled from a multivariate normal distribution with standard deviation that is a fixed fraction of the mean (this fraction is $0.02$ for demand and  $0.15$ for wind). Moreover, a monthly trend in demand and wind is governing the mean profiles \cite{ouammi2010monthly}; the choice of month is drawn uniformly. For more details on the distributions used to generate samples in this work please refer to Appendix~\ref{sec:transition_model}.
	
	For the sake of sampling daily network topology, we consider the list of candidate planned outages allegedly requested by the TSO as given in Fig.~\ref{fig:case96_modifications}. The list consists of $7$ outages per each of the three zones of RTS96, plus $3$ interconnection outages. A straight-forward implementation of  UCNN  would have required a huge dataset that is $O(2^{3\times7+3})$ due to the possible outage combinations. Therefore, in this work, for the considered outages we assume that each of the three ``zone operators'' receives his exclusive time allocation throughout the year to conduct his $7$ outages. By this, we do away with the exponential dependence of UCNN's complexity in the number of zones, i.e., reduce the $O(2^{3\times7+3})$ training set size to $O(3\times2^{7+3})$.

	Each outage combination sample, i.e., topology $\text{top}_{\text{d.a}}(x)$, is drawn uniformly from the set of outage combinations. As for available generators and costs parameters, the whole study assumes those remain fixed.	
	
\begin{figure}
		\centering
		\includegraphics[scale=0.31]{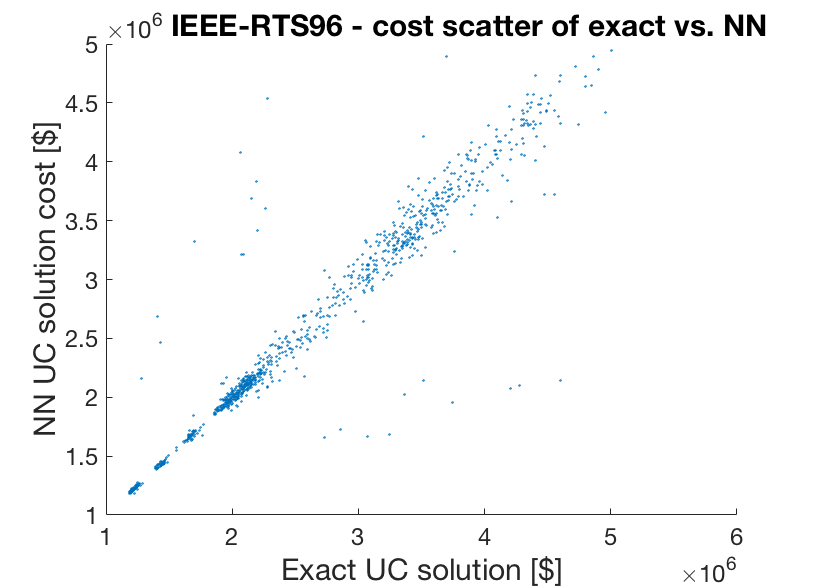}
		\caption{A scatter plot, comparing accurate UC costs  $C^*_p(x_\text{test})$ and corresponding UCNN predicted costs $\hat{C}^{*}_p(x_\text{test})$, for all $x_\text{test} \in {\mathcal{X}}_{\text{test}}$. Average relative error measures $3.6\%$, while linear correlation coefficient is $0.96$.  } \label{fig:UCNN_costs}
	\end{figure}

	 A scatter plot of accurate UC costs  vs. UCNN costs is presented in Fig. \ref{fig:UCNN_costs}. The form of small clusters is obtained since several season-dependent daily mean demand-wind forecast profiles are used. During summer demand is low, and generation cost is relatively low. The months of this season correspond to the three small clusters of low UC costs. As a result, these improve linear correlation compared to when costs are high. The relative error measure is more robust to this effect, due to its denominator as given in Subsection~\ref{sec:evaluation methodolohy}. Overall, the resulting accuracy reports to be high: average relative error measures $3.6\%$, while linear correlation coefficient is $0.96$.
	
\begin{figure}
		
\centering
				\includegraphics[scale=0.31]{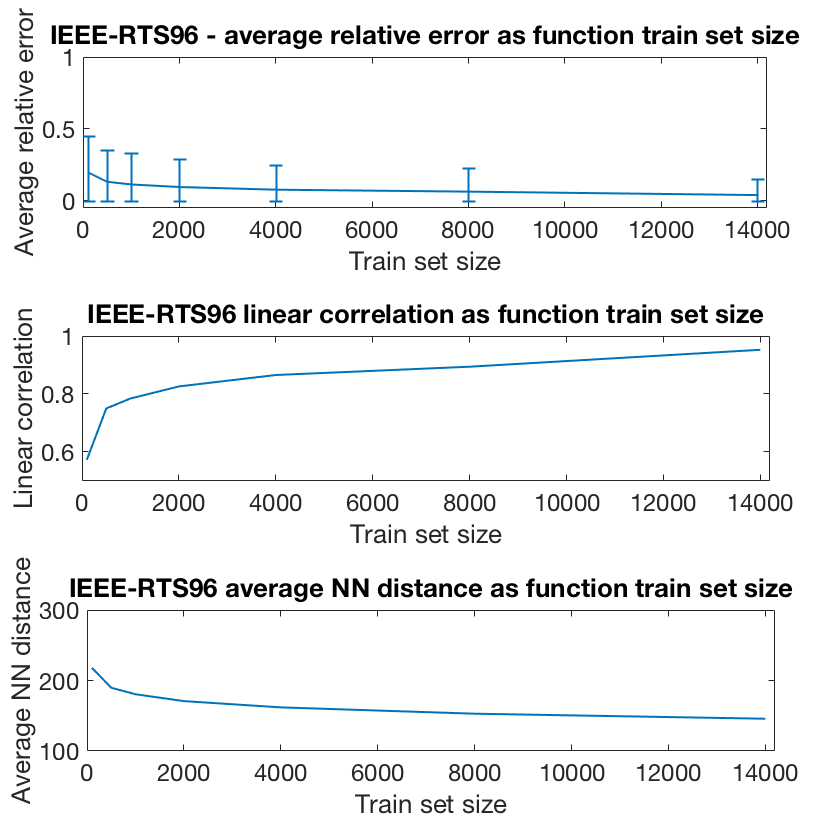}

			\label{fig:Ng2}
		
		\caption[training set size examination]{Average relative error, linear correlation coefficient, and average distance to nearest neighbor, as a function of $|{\mathcal{X}}_{\text{train}}|$. The larger the training set, the better UCNN algorithm performs.} \label{fig:train_size}
	\end{figure}

	An imperative question is how big should the training set be. Given some fixed level of accuracy to be achieved, there is an obvious dependence of the required size of ${\mathcal{X}}_{\text{train}}$ on the dimensionality of samples and their variance. The dimensionality is the total number of load buses, wind generators, and candidate outages. The variance is based on common values from the literature, as brought in the opening of this section. Therefore, the larger the considered power system model and the more outages investigated, the larger the training set should be. We leave the detailed analysis discovering a mathematical relationship between the two as an open question. Nevertheless, we now present an empirical examination of the level of approximation accuracy as a function of the training set size. Fig. \ref{fig:train_size} contains three plots:  average relative error,  correlation, and density as a function of $|{\mathcal{X}}_{\text{train}}|$. The third metric, density, is defined to be the average distance (as defined in \eqref{eq:distance}) of $x_\text{test} \in  {\mathcal{X}}_{\text{test}}$ from its nearest neighbor in ${\mathcal{X}}_{\text{train}}$. The results demonstrate errors that can be considered tolerable at already small sizes. A training set  of full size, $14K$, is sufficient for obtaining relative error and correlation of $3.6\%$ and $0.96$, using the examined setting. As the training set grows, the smaller the average distance is from nearest neighbors, allowing for a better approximation.

	Next, we discuss the runtime improvement aspect. We compare the average runtime of solving an accurate UC program\footnote{Overall runtime includes MILP modeling time, roughly taking 30-40\% of the calculation time.} and obtaining a single UC solution when using UCNN. While accurate UC spans over 81 seconds on average, UCNN runtime is 0.31 seconds, spent on searching for the nearest neighbor of a sample. This two orders of magnitude reduction is significant, and as shown in \cite{dalal2016distributed}, is, in fact, a turning point for making long-term assessment tractable. This gain comes at the price of the initial UCNN training time, which, in the experimental setup described here, is roughly two days (but can be improved with more parallelized hardware). Other parts of the simulation, e.g., probabilistic scenario generation, are in the order of a few milliseconds and thus are negligible. 
	
	\subsection{Additional investigation and further experiments}
	\label{sec:further}
	Our investigation of UCNN was extended in several directions, which, due to lack of space, we enclose in short. First, we experimented on a second test-case, IEEE-RTS79, that was modified in spirit that is similar to the modifications described in Section \ref{sec:UCNN_experiments}. The results exhibit practically identical behavior for both networks in all simulations. To achieve the same low-error results as reported in Figures \ref{fig:UCNN_costs} and \ref{fig:train_size}, a training set size that is roughly 2.5 smaller was enough for IEEE-RTS79 compared to IEEE-RTS96. This supports our claim that larger networks require more pre-computation.
	 
	Second, we compared our method to N-1 secured UC; the difference is primarily in runtime. Adding N-1 security constraints results in accurate solution times that are an order of magnitude larger than the non N-1 case, whereas the UCNN runtime remains unchanged as expected. We therefore achieve runtime gain of three orders of magnitude with UCNN, rendering the method even stronger for that case.
	 
	Third, in addition to using cost as the sole classification accuracy criterion, we consider a notion of reliability in terms of resiliency to N-1 events; i.e., the fraction of single-line outages on top of a given topology, for which ACPF convergence is obtained. When comparing these values for accurate vs. UCNN solutions, we again witness strong correlation and low relative error.
	
	Lastly, in other recent work \cite{dalal2018chance}, we combine UCNN as part of an outage scheduling scenario assessment mechanism, used for finding optimal scheduling plans. There, we show how UCNN accurately predicts several cost and reliability metrics, and helps reduce overall outage scheduling evaluation and optimization runtimes by several orders of magnitude.

	\section{Conclusion}
	In this work, we argue that at times, the accuracy vs. runtime trade-off is not to be resolved by solely focusing on the former. We harness the power of machine learning 
	and present the notion of a proxy -- a module that approximates short-term decision making outcomes in a hierarchical setting, thus facilitating tractable assessment methodologies. 	
	
	The potential overall gain in CPU time is the fundamental advantage of this method when used in the context of long-term assessment/control applications. This gain is essentially constituted by the ratio between the overall number of UC programs being solved in the process of long-term assessment, and the size required for UCNN's training set. As shown in \cite{dalal2016distributed}, the number of UC programs solved for assessment can be orders of magnitude larger than the training set sizes used in Section \ref{sec:UCNN_experiments}. This potential CPU-time speed-up is elevated even further when long-term planning is performed, and multiple iterations of assessments are conducted. Moreover, when implemented using efficient data-structures such as KD-Trees \cite{samet1990design}, computational complexity for finding a NN is sub-linear, eliminating the need of iterating over all data. Additionally, the merits of our method hold even with small training sets: only $8\%$ relative error is witnessed for a training set of $1000$ samples in the case of IEEE-RTS96 (see Fig.~\ref{fig:train_size}). 

	Potential further research can tackle the metric-learning problem, discovering metrics induced by the classification problem at hand. Our belief is that such an approach could not only improve prediction accuracy but also bear insights regarding the importance of different components of power networks in terms of cost and reliability.  An additional direction is to use UCNN as a warm-start strategy to the accurate UC optimization problem, where the approximate solution is fed to the solver as an initial guess. This enhancement can speed-up accurate UC solutions and can be of interest when approximations are not satisfactory.
	
	\bibliographystyle{IEEEtran}
	\bibliography{IEEEabrv,PES16}
	
	\appendices
	\section{Daily conditions distributions}  \label{sec:transition_model}
	Generation of the training and test sets involves sampling UC inputs $\{x_i\}_{i=1}^n$ from distribution $\mathbb{P}_X(x)$. Our sampling technique is based on the following factorization of the random vector $x$: daily wind power $W_{\text{d.a}}$ and daily load $D_{\text{d.a}}$ are statistically independent conditioned on the month of the year (which is drawn uniformly first), whereas daily network topology $\text{top}_{\text{d.a}}$ is independent of them both. Each independent component is thus sampled as follows.
	\subsubsection{Wind power distribution}
	Wind generation capacities are taken from \cite{UW_website}, along with their daily mean profile. 
	The wind process mean $\mu_w(t)$ is  obtained from the formula \[\mu_w(t) = \mu_w(t_D)\cdot p_{w,\text{monthly}}(t_M),\] where $\mu_w(t_D) \in \mathbb{R}_+^{n_w}$ is the daily wind mean profile at time-of-day $t_D$, and $p_{w,\text{monthly}}(t_M) \in [0,1]$ is the monthly wind intensity relative to its peak at month $t_M$. 
	
	The hourly wind generation $W_t$ is multivariate normal: \[W_t \sim \mathcal{N}\left(\mu_w(t),diag((p_{w,\sigma}\cdot \mu_w(t))^2)\right),\] where $p_{w,\sigma} \in [0,1]$ is a constant ($=0.15$) that multiplies the mean $\mu_w(t)$, to obtain a standard deviation that is a fixed fraction of the mean; $diag(x)$ is a square diagonal matrix, with the elements of $x$ as its diagonal, assuming wind generators to be uncorrelated; and  $W_t$ is truncated to stay in the range between $0$ and the generator's capacity.
	\subsubsection{Load distribution}
	Hourly load $D_t$ is assumed to follow the same normal distribution as the wind, with the same formula containing peak loads and daily profiles for each bus $\mu_d(t_D) \in \mathbb{R}_+^{n_b}$ with values taken from \cite{UW_website}.
	Fraction of mean for standard deviation is set to be $p_{d,\sigma}=0.02$.
	\subsubsection{Outage distribution} 
	Section~\ref{sec:UCNN_experiments} lists the choice of transmission lines where outages are considered. Sampling of daily topology $\text{top}_{\text{d.a}}$ is done  uniformly out of the combinatorial outage set.

	\section{Exact unit commitment formulation} \label{sec:unit_commitment}
	The accurate unit-commitment problem formulation is 
	\small
	\begin{subequations}  \label{eq:unit_commitment}
			\begin{align}
		& u_p^* = \argmin_{u_p \in \mathcal{U}_p(u_m, y_s)}   C_{p}(u_m,u_p) = \argmin_{\alpha,\Theta,P_{g,t},WC,LS} \nonumber \\
		& \sum_{t=1}^{T{d.a}} \left[\sum_{i=1}^{n^g_d}\left(\alpha_{t}^if_P^i(P_{g,t}^i) + \alpha_{t}^i(1-\alpha_{t-1}^i)SU_i(t_{\text{off}}^i(\alpha,t) \right)\right.   \nonumber \\
		&\quad \quad \quad + \left. \sum_{iw=1}^{n^g_w}WC_{t}^{iw}\cdot C_{WC} + \sum_{ib=1}^{n^b}LS_{t}^{ib}\cdot VOLL \right], \label{eq:objective}\\
		&\text{subject\ to}  \\
		& g_{P,t}^l(\Theta^l,\alpha,P_g)=~B_{\text{bus}}^l \Theta_{t}^l + P_{BUS,\text{shift}}^l + {D}_{d.a,t} \label{eq:power_balance} \\
		& ~~+G_{sh} - LS_{t}  - ({W}_{d.a,t}-WC_{t})  - C_g (\alpha_{t}.* P_{g,t}) = 0, \nonumber\\
		& h_{f,t}^l(\Theta_{t}^l) =  B_f^l \Theta_{t}^l + P_{f,\text{shift}}^l - F_{max}^l \leq 0, \\
		& h_{t,t}^l(\Theta_{t}^l) = B_f^l \Theta_{t}^l - P_{f,\text{shift}}^l - F_{max}^l \leq 0, \label{eq:to_line_limits} \\
		& \theta_i^{\text{ref}} \leq \theta_{i,t}^l \leq \theta_i^{\text{ref}}  \quad i \in {\cal I}_{\text{ref}}, \label{eq:angle_limits}\\
		& \alpha_{t}^iP_g^{i,\text{min}} \leq P_{g,t}^i \leq \alpha_{t}^iP_g^{i,\text{max}} \quad  i=1,\dots,n^g_d,\\
		& 0 \leq WC_{t}^{iw} \leq {W}_{d.a,t}^{iw} \quad  iw=1,\dots,n^g_w,\\
		& 0 \leq LS_{t}^{ib} \leq {D}_{d.a,t}^{ib} \quad  ib=1,\dots,n^b, \label{eq:LS_limit}\\
		& t_{\text{off}}^i(\alpha,t) \geq  t_{\text{down}}^i \quad  i=1,\dots,n^g_d, \label{eq:min_down}\\
		& t_{\text{on}}^i(\alpha,t) \geq  t_{\text{up}}^i\quad   i=1,\dots,n^g_d, \label{eq:min_up}\\
		&l=0,1,\dots,n^l_t, \label{eq:n-1 constraint}\\
		&t=1,\dots,T_\text{d.a}.
		\end{align} 
	\end{subequations}
	\normalsize
	Formulation \eqref{eq:unit_commitment} generally supports ensuring the N-1 security criterion via \eqref{eq:n-1 constraint}. The N-0 case is obtained by replacing \eqref{eq:n-1 constraint} with $l=0.$ The formulation's components are explained as follows.
	
	\begin{itemize}
		\item $l$ denotes index of a transmission line that is offline. $l=0$ denotes all lines are connected and online. lines undergoing an outage are excluded from $n^l_t.$ 
		\item $\alpha \in \{0,1\}^{n_d \times T_\text{d.a}}$ denotes commitment (on/off) status of all dispatchable generators at all time-steps.
		\item $\Theta \in [-\pi,\pi]^{ n_b \times (n_l+1) \times T_\text{d.a}}$ denotes  voltage angle vectors for the N-1 network layouts at all time steps.
		\item $P_g \in \mathbb{R}_+^{n_d \times T_\text{d.a}}, WC \in \mathbb{R}_+^{n_w \times T_\text{d.a}}, LS \in \mathbb{R}_+^{n_b \times T_\text{d.a}}$ denote dispatchable generation, wind curtailment and load shedding decision vectors, with $f_P, C_{WC}, VOLL$ being their corresponding prices.
		\item $t_{\text{down}}^i, t_{\text{up}}^i$ denote minimal up and down time limits for generator $i$, after it had been off/on for $t_{\text{off}}^i$/$t_{\text{on}}^i;$ the latter are functions of $\alpha$ and $t$, as depicted in $\eqref{eq:objective}.$
		\item $SU_i(t_{\text{off}}^i(\alpha,t))$ denotes start-up cost of dispatchable generator $i$ after it had been off for $t_{\text{off}}^i$ time-steps.
		\item $g_{P,t}^l(\Theta^l,\alpha,P_g)$ denotes the overall power balance equation for line $l$ being offline.
		\item $B_{\text{bus}},P_{BUS,\text{shift}}$ denote nodal real power injection linear  coefficients.
		\item $B_f,P_{f,\text{shift}}$ denote linear coefficients of the branch flows at the \emph{from} ends of each branch (equal minus of the \emph{to} ends, due to the lossless assumption).
		\item $G_{sh}$ denotes a vector of real power consumed by shunt elements.
		\item $C_g$ denotes generator-to-bus connection matrix, where $(\alpha_{t}.* P_g)$ denotes the dot-product of the two vectors.
		\item $F_{max}$ denotes line flow limits.
		\item ${\cal I}_{\text{ref}}$ denotes the set of indices of reference buses, with $\theta_i^{\text{ref}}$ being the reference voltage angle.
		\item $P_g^{i,\text{min}},P_g^{i,\text{max}}$ denote minimal and maximal power outputs of generator $i$.
	\end{itemize}

	Furthermore,
	\begin{itemize}
		\item \eqref{eq:power_balance}-\eqref{eq:to_line_limits} ensure load balance and network topology constraints;
		\item \eqref{eq:angle_limits}-\eqref{eq:LS_limit} restrict the decision variables to stay within boundaries. Namely, voltage angle limits, generator minimal and maximal power output range, wind curtailment and load shedding limits; and
		\item \eqref{eq:min_down}-\eqref{eq:min_up} bind the different time steps to follow generator minimal up and down time thermal limits.
	\end{itemize}


%
%
%
%

	
	

\end{document}